\documentclass[runningheads]{llncs}
\usepackage{graphicx}
\usepackage{hyperref}

\usepackage{times}
\usepackage{latexsym}
\usepackage{parcolumns}
\usepackage{lipsum}

\usepackage{caption}
\usepackage{graphicx}
\usepackage{booktabs}
\usepackage{comment}

\usepackage{blindtext}
\usepackage{wrapfig}

\usepackage[T1]{fontenc}
\usepackage[utf8]{inputenc}
\usepackage{babel}
\usepackage[font=small,labelfont=bf]{caption}

\usepackage[T1]{fontenc}

\usepackage[utf8]{inputenc}

\usepackage{microtype}

\usepackage{amssymb}
\usepackage{amsmath, amsfonts}
\usepackage{makecell}
\usepackage{siunitx}
\usepackage{color}
\usepackage{arydshln}
\usepackage{graphicx}
\usepackage{tabularx}
\usepackage{booktabs}
\usepackage{multirow}
\usepackage{multicol}
\usepackage{pifont}

\DeclareMathOperator*{\argmax}{arg\,max}

\newcolumntype{P}[1]{>{\centering\arraybackslash}p{#1}}

\usepackage{subcaption}

\makeatletter
\newcommand{\ssymbol}[1]{^{\@fnsymbol{#1}}}
\makeatother

\makeatletter
\def\@seccntformat#1{\@ifundefined{#1@cntformat}%
   {\csname the#1\endcsname\quad}  
   {\csname #1@cntformat\endcsname}
}
\let\oldappendix\appendix 
\renewcommand\appendix{%
    \oldappendix
    \newcommand{\section@cntformat}{\appendixname~\thesection\quad}
}
\makeatother

\begin{document}

\graphicspath{{figs/}}
\title{Task Conditioned BERT \\for Joint Intent Detection and Slot-filling}
\author{Diogo Tavares\inst{1}
Pedro Azevedo\inst{2}
David Semedo\inst{1}
Ricardo Sousa\inst{2}
João Magalhães\inst{1}
}
\authorrunning{Tavares, D. et al.}

\institute{Universidade NOVA de Lisboa \and
Farfetch\\
\email{dc.tavares@campus.fct.unl.pt, \{df.semedo, jm.magalhaes\}@fct.unl.pt}}
\maketitle              
\begin{abstract}
Dialogue systems need to deal with the unpredictability of user intents to track dialogue state and the heterogeneity of slots to understand user preferences.
In this paper we investigate the hypothesis that solving these challenges as one unified model will allow the transfer of parameter support data across the different tasks.
The proposed principled model is based on a Transformer encoder, trained on multiple tasks, and leveraged by a rich input that conditions the model on the target inferences.
Conditioning the Transformer encoder on multiple target inferences over the same corpus, i.e., intent and multiple slot types, allows learning richer language interactions than a single-task model would be able to.
In fact, experimental results demonstrate that conditioning the model on an increasing number of dialogue inference tasks leads to improved results: on the MultiWOZ dataset, the joint intent and slot detection can be improved by 3.2\% by conditioning on intent, 10.8\% by conditioning on slot and 14.4\% by conditioning on both intent and slots. 
Moreover, on real conversations with Farfetch costumers, the proposed conditioned BERT can achieve high joint-goal and intent detection performance throughout a dialogue.

\keywords{Dialogue State Tracking\and Intent detection \and Slot filling \and BERT.}
\end{abstract}

\def\tokens #1{\fcolorbox{black}{white}{\rule{0pt}{4pt}\rule{0pt}{0pt}#1}}

\definecolor{cyan2}{rgb}{.4,.8,0.9}
\def\cls {\fcolorbox{black}{cyan}{\rule{0pt}{4pt}\rule{0pt}{0pt}[CLS]}}
\def\sep {\fcolorbox{black}{cyan}{\rule{0pt}{4pt}\rule{0pt}{0pt}[SEP]}}
\def\usr {\fcolorbox{black}{cyan}{\rule{0pt}{4pt}\rule{0pt}{0pt}[USR]}}
\def\sys {\fcolorbox{black}{cyan}{\rule{0pt}{4pt}\rule{0pt}{0pt}[SYS]}}

\definecolor{cyan}{rgb}{.4,.9,0.9}
\def\intent {\fcolorbox{black}{green}{\rule{0pt}{4pt}\rule{0pt}{0pt}[INTENT]}}

\definecolor{green2}{rgb}{.9,.9,0.4}
\def\slotage {\fcolorbox{black}{orange}{\rule{0pt}{4pt}\rule{0pt}{0pt}[SLOT-AGE]}}

\definecolor{red2}{rgb}{.8,.8,0.4}
\def\slotgender {\fcolorbox{black}{lime}{\rule{0pt}{4pt}\rule{0pt}{0pt}[SLOT-GENDER]}}

\section{Introduction}
Conversational assistants need to explicitly maintain information about user goals by tracking the user intent and storing a set of \textit{slot-value pairs}. 
This is critical to ensure the smoothness of user-agent interaction leading to frustration-free outcomes.
Both dialogue state and slot values can be used as a way to provide a general initial product suggestion~\cite{shoptalk}, before more fine-grained attributes are requested by the system.
Hence, keeping the dialogue agent up-to-date with user's perception of the current conversation is a critical, yet, non-trivial task~\cite{simpletod}.

Algorithms that support more natural conversations need to tackle complex phrasal constructions~\cite{bertdst} and dialogue contextual information~\cite{tripy}.
Each user utterance conveys multiple and intertwined hints leading to very rich language structures and possible co-references to the dialogue history. 

Recent approaches~\cite{bertdst,purelytransformerdst,tripy,simpletod}, explored the Transformer model in this context and leveraged the attention mechanisms to tackle the above challenges.
A common practice is to use the control token to detect intent~\cite{bertjointclassslotfill,todbert,wu-etal-2020-slotrefine} or presence of a slot span~\cite{unknownslotvaluesdst,bertdst,schemaguided}.

Recent works extend the Transformer with new heads~\cite{bertjointclassslotfill,todbert,wu-etal-2020-slotrefine}, tackling both intent detection and slot filling in a multi-task setting.
While these works capture the dependencies between intent detection and slot-filling, all the inferences are solely conditioned on the dialogue utterances, without accounting for each target inference task.

Our research hypothesis is that jointly learning dialogue inference tasks while conditioning the Transformer on the aforementioned dialogue state-tracking (DST) tasks, will lead to more precise joint-inferences of user intent and slot filling, i.e., more accurate dialogue state inferences.
This hypothesis is supported by the way BERT~\cite{devlin-etal-bert} attends to different tokens~\cite{bertlookat} -- the [CLS] token, retaining a global sequence embedding, can leverage a number of language tasks~\cite{devlin-etal-bert}, by functioning as an attention hub, contextualizing the whole input sequence. 
Extra special attention hub tokens can then be added and learned through fine-tuning. 
Hence, we argue that introducing new task-specific tokens, acting as task-specific attention hubs, alongside Transformer heads, could allow for the introduction of additional domain-specific operations. 
We argue that these empirical observations are all rooted on the same principle: \textit{when the Transformer encoder is conditioned on the target task, the self-attention mechanism across all layers becomes aware of the target inference operation}. Thus, the conditioning input can steer the inferences across all layers. This forms the base assumption of our work.

In the following section we discuss the related work. In sections~\ref{sec:formalization} and~\ref{sec:taskconditioned} we describe the proposed approach. Section~\ref{sec:eval} presents and discuss experimental results.

\section{Related work}

\textbf{Dialogue State Tracking (DST)} refers to the act of maintaining a set of user goals or preferred attributes by performing slot-filling in task-oriented dialogues, which can be either single or multi-domain.
Span-based slot-filling approaches have been widely explored with promising results, as seen in ~\cite{unknownslotvaluesdst}, \cite{bertdst}, \cite{schemaguided}, with the first employing RNN encoding and the latter two using a BERT-based encoder. Extracting spans may sometimes be sufficient to attain good performance, but, in open-ended dialogues, may prove insufficient when facing values implicitly mentioned by the user or values which refer to previously filled slots.
To remedy this, work towards introducing other types of information has been developed, maintaining the same BERT encoder setup. \cite{tripy} proposed to directly refer the previously made slot assignments or system suggestions, depending on the output of the slot-gate, which is extended so as to perform a more fine-grained classification. Other approaches, such as~\cite{findorclassifydst}, make use of predefined ontologies when slots are considered categorical. While non-categorical slots are classified by detecting relevant spans in the dialogue, categorical slots use a fixed BERT model to encode all possible slot key-value combinations in the ontology, and use cosine similarity matching with the [CLS] token output of both BERT instances. While this work is similar in spirit to ours, we directly adapt BERT-DST~\cite{bertdst} to develop our models, as was previously attempted by~\cite{tripy}.

BERT-DST~\cite{bertdst} classifies each slot independently from one another in two steps: first, using BERT’s [CLS] token embeddings, it classifies whether a slot is or is not present in the utterances, or whether the user expressed no interest in its value; referred to as a \textbf{slot-gate}. Second, for each slot where the slot-gate output is positive, using the embedding of each token, attempts to extract the dialogue span in which its value is mentioned.

\vspace{3mm}
\textbf{Intent Detection} requires analyzing a user utterance and classifying it, as a whole, given a set of possible user intents.
Transformer encoder-based approaches are especially adept at this task, performing the classification step using sentence embeddings.
Intent detection data is limited in task-oriented datasets, and most approaches \cite{bertjointclassslotfill,improvingslotfilling,stackpropagation} focus on single-utterance queries for voice assistants \cite{atis,snips}, forgoing multi-turn interactions.

\vspace{3mm}
Recently developed \textbf{DST datasets}, such as \cite{schemaguided,multiwoz2.2}, have attempted to account for the fact that real-world systems will contain categorical and non-categorical slots. Alongside this notion, they also push the relevance of intent detection, with \cite{schemaguided} supplying intent annotations and \cite{multiwoz2.2}, an update to MultiWOZ~\cite{multiwoz}, updating the annotation set with user intent annotations.

\section{Proposed Model}
\label{sec:formalization}
Slot-filling and intent detection are natural language processing tasks associated to the understanding of a sequence $\mathcal{D}=\{(u_1, a_1), \mathellipsis , (u_{T}, a_{T})\}$, of $T$ dialogue turns, where each turn $i$ is represented by a tuple $(u_i, a_i)$ composed of user and system utterances, respectively.
First, given the user utterance $u_{T+1}$ and a set of $M$ possible intents $\mathcal{I}=\{I_1, \mathellipsis ,I_{M}\}$,
our goal is to infer the correct intent $I_m$ of the user utterance. 
Second, given all dialogue utterances up to turn $T$ and a set of $N$ slot-keys $\mathcal{S}=\{s_1, \mathellipsis , s_{N}\}$, 
the goal is to assign a slot-value $v \in \{v_1, \ldots,  v_i, \ldots \}$ to every slot-key $s_k$ which was, explicitly or otherwise, accepted or suggested by the user in the turns present in $\mathcal{D}$. A slot-value can be anything from a \textit{hotel location} to the \textit{number of people} in a restaurant reservation.
The act of maintaining all relevant slot key-value pairs in a dialogue $\mathcal{D}$ is referred to as \textit{Dialogue State Tracking} (DST). 

\subsection{Dialogue Task Conditioned Encoder}
\label{sec:taskconditioned}
Conditioning the Transformer encoder on dialogue data can be achieved by considering the entire sequence of dialogue utterances. We can consider the independent probabilities of user intent $p(I_m | u_T, \mathcal{H}_c)$ and slot key-value $p(s_k = v_i | u_T, \mathcal{H}_c)$ where $u_T$ stands for the current user utterance, and $\mathcal{H}_c=\{u_{T-c},\ldots,u_{T-1}\}$ is the set of past dialogue utterances.
Alternatively to the independent modes, the joint-inferences of intent and slot filling is an explicitly dependency-based model, $p(I_m, s_k = v_i | u_T, \mathcal{H}_c)$ where the joint inference is, again, conditioned on the dialogue history $\mathcal{H}_c$. We extend these variables and investigate how different conditioning assumptions affect the Transformer inference performance for joint slot-filling and intent detection.
In practice, we enrich the conditional probability with dialogue task information $DT$,\begin{equation}
    p(I_m, s_k = v_i |u_T, \mathcal{H}_c, DT),
\end{equation}
which brings a series of advantages to Transformer-based implementations of the above model.

\subsection{Dialogue Task Conditioning}
Large Transformer models~\cite{todbert,simpletod} are able to singlehandedly model complex tasks within dialogues, such as next sentence prediction, intent detection, and ontology-based slot-filling. 
Even though intent detection in TOD-BERT~\cite{todbert} is performed by leveraging the [CLS] token, both SimpleTOD~\cite{simpletod} and TOD-BERT prepend user and assistant utterances with special tokens that denote the speaker. 
In DST, user and assistant turns should be attended differently: in order to perform slot-filling on a slot key, the user must either state it (explicitly or otherwise) or agree with an assistant suggestion. 
The aforementioned tokens can \textit{condition} the Transformer into performing slot-filling appropriately in each situation. SimpleTOD~\cite{simpletod} further makes use of tokens to delineate the start and end of each dialogue subtask, such as slot-filling and response generation.

\begin{figure*}[t]
    \centering
    \includegraphics[width=0.9\linewidth]{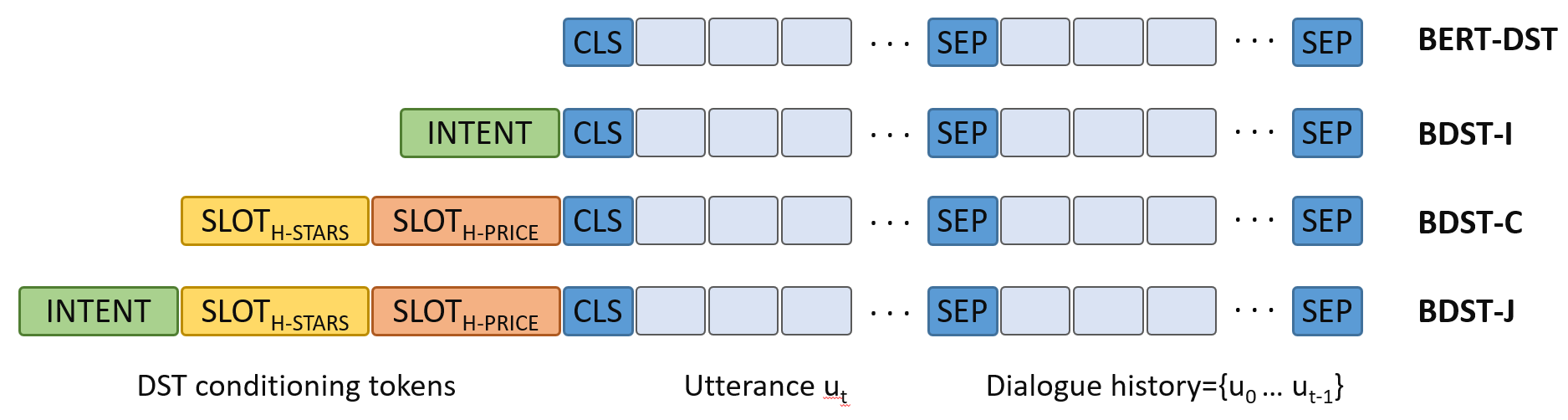}
    \caption{The dialogue target task is explicitly passed to the encoder to condition its inferences.}
    \label{fig:input-formatting}
    \vspace{-5mm}
\end{figure*}

Hence, in light of what we know~\cite{todbert} regarding special token usage on vanilla BERT ([CLS], [SEP]) and pre-trained TOD systems (utterance source tokens, subtask delineation), we pass dialogue specific tokens to the encoder to condition its inference operations (Figure~\ref{fig:input-formatting}).
Each one of these dialogue specific tokens is then fine-tuned on the corresponding target inference tasks.
This is extremely important since now, all encoder layers will have explicit information regarding the required output task.

\subsection{BERT-DST: Span slots}
\label{sec:bertdstupdates}
First, we build on the BERT-DST~\cite{bertdst} model and leverage the fact that BERT overly attends to special tokens~\cite{bertlookat}.
This baseline model uses the standard input formatting~\cite{devlin-etal-bert} (first row of Figure~\ref{fig:input-formatting}), 
where each input token is mapped to an $h$  dimensional internal representation. The output $\mathbf{O}\in \mathbb{R}^{L\times h}$ comprises contextualized embedding representations of the input tokens.

\begin{figure*}[t]
    \centering
    \includegraphics[width=0.9\linewidth]{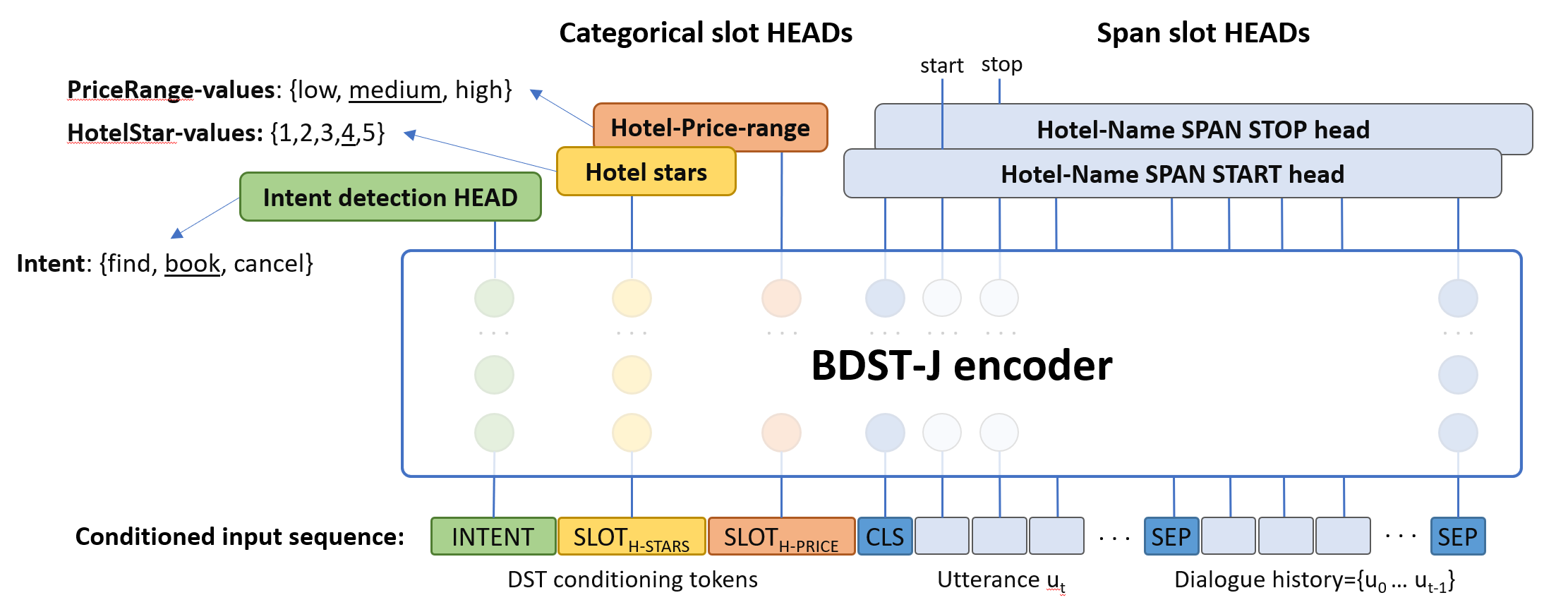}
    \caption{The BDST-J architecture explicitly conditions the dialogue state inference operations in an end-to-end fashion over the intent and domain-slots.}
    \label{fig:bert_dst_architecture}
    \vspace{-5mm}
\end{figure*}

As previously described, the [CLS] token feeds the slot-gate \textit{softmax} layer, and the slot values are extracted using a span-based approach over $\mathcal{D}$.
The span detection is implemented as two classification layers, one for the \textit{span-start} and one for the \textit{span-end}, see Figure~\ref{fig:bert_dst_architecture}. All these layers are trained under a common loss function
\begin{equation}
\begin{split}
    \mathcal{L}_{slot} = &\alpha \cdot \mathcal{L}_{slot\_gate} + \frac{1- \alpha}{2} \cdot \Large(\mathcal{L}_{span\_start} + \mathcal{L}_{span\_end} \large),
\end{split}
\label{eq:bert-dstloss}
\end{equation}
, a convex combination parameterized by $\alpha$.

\subsection{BDST-I: Intent Detection}
Our first take towards conditioning the Transformer encoder in the target inference task is to introduce an [INTENT] token to the sequence input.
This new token embedding is used by a linear classification layer head to detect the intent. 

Introducing the aforementioned token is feasible as both tasks are inherently related---in fact, recent DST approaches \cite{schemaguided} attempt to consolidate intent detection and slot-filling within the same model. We also argue that slot classification is inherently coupled with the current user intent. When users \textit{intend} to, for instance, request hotel information, it is more likely that they would mention the \textit{number of people} than also request a \textit{restaurant location} in the same turn. This is also shown by a strong Cramer's V correlation~\cite{coefficients} between utterances of a specific intent and mentioned slots, on all considered datasets (discussed in Section \ref{sec:token-datasets}). Specifically, the MultiWOZ and Farfetch-Costumers datasets both exhibit a 0.62, Farfetch-Sim 0.53, and Sim-R with 1.

We fine-tune BDST-I to both slot-filling and intent detection, adding $\beta\cdot\mathcal{L}_{intent}$ to the BERT-DST loss function (eq. \ref{eq:bert-dstloss}), with $\mathcal{L}_{intent}$ as the cross entropy loss for the intent prediction target, and $\beta$ is a convex combination constants:

\begin{equation}
     \mathcal{L}_{BDST-I} = \beta \cdot \mathcal{L}_{intent} + (1-\beta)\cdot\mathcal{L}_{slot}
\end{equation}

The embedding weights of the [INTENT] token are initialized with the [CLS] weights and are then fine-tuned to the intent detection task.
$\beta$ was determined experimentally on the validation set.

\subsection{BDST-C: Categorical slots}
The search for the presence of slots is usually focused on the ones that make sense for the current dialogue stage---in real world scenarios, it is not plausible to search for all slots in all dialogue stages. Thus, for each categorical slot that we wish to detect, we introduce a slot-specific input token, each initialized with random embeddings, signaling we need to perform inference on each mentioned slot.
The BERT model input is shown in Figure~\ref{fig:input-formatting}: assuming \textit{hotel-stars} and \textit{hotel-price} as the categorical slots in the domain.
In such cases, given a categorical slot [cs], whose possible values are in $V_{[cs]}$, and the corresponding token $BERT_{cs}$, the slot value is determined by a classifier head,
\begin{equation}
    \argmax_{V_{cs}} \ W_{cs} \cdot BERT_{cs} + b_{cs}
\end{equation}
where $V_{cs}$ is the set of all possible values for slot key $[cs]$ in the domain ontology.
Note that in domains without categorical slots, the model input is the same as vanilla BERT-DST.

BDST-C uses a different classification strategy depending on the slot type, so special considerations must be taken. We use a weighted sum for the loss, as follows:
\begin{equation}
     \mathcal{L}_{BDST-C} = \beta \cdot \mathcal{L}_{cat} + (1-\beta) \cdot \mathcal{L}_{slot}
\end{equation}

Following the assumption that each slot is of equal importance to the final result, we fix $\beta$ to $(\#categorical\;slots)/(\#total\;slots)$.

\subsection{BDST-J: Joint Intent and Multiple-Slots}
As previously mentioned, both extensions attempt to exploit BERT being capable of assigning operations to special tokens. Similarly to how [CLS] is known to contain an aggregate sequence representation for NSP, it is easy to see how an [INTENT] token could also contain an aggregate representation based on all the possible intents. The same rationale applies to the extra categorical tokens, potentially containing sentence-level representations weighted on the semantic classification of specific slot-keys.
Hence, we generalize the above approaches and introduce a fully flexible input sequence for the joint task, BDST-J, Figure~\ref{fig:input-formatting}.
It follows that, when training BDST-J, the loss function is:
\begin{equation}
\mathcal{L}_{BDST-J} = \alpha \cdot \mathcal{L}_{BDST-I} + (1-\alpha) \cdot \mathcal{L}_{BDST-C}
\end{equation}
All parameters are determined on the validation set.

\section{Evaluation}
\label{sec:eval}
In this section we evaluate the vanilla BERT-DST model, BDST-I, BDST-C, and BDST-J on Sim-M, Sim-R, MultiWOZ 2.2 benchmarks, and on the Farfetch dataset, with real testers.
All the baselines we tested are encoder-only architectures and have a similar number of parameters for a fair comparison, with the exception of the low-parameter TRADE-DST~\cite{tradedst}. Other architectures require more training time and are more complex to deploy. 

\subsection{Datasets}
\label{sec:token-datasets}
\vspace{2mm}
\noindent\textbf{M2M (Sim-M + Sim-R).}
Sim-R and Sim-M~\cite{SIM}, respectively focusing on the restaurant and movie ticket domains, use crowdsourced paraphrasing of template utterances to simulate both user and agent. \textit{All slots are non-categorical}, which biases the dialogue towards simple and direct conversations where slot values are \textit{always} explicit in utterances. Dialogues are also noiseless, which may not reflect some of the challenges of an in production, robust DST system. Both datasets have a high proportion of out of vocabulary values, meaning that several test set slot values are absent during training. These values are contained in the \textit{restaurant\_name} and \textit{movie} slots. Sim-R contains coarse-grained intent detection, with two possible intent values: \textit{find} and \textit{reserve restaurant}. Compared to other datasets used in this work, the amount of dialogues is relatively low---to perform well on M2M, models must develop a robust understanding of the semantics of slot-filling with sparse data. 

\vspace{2mm}
\noindent\textbf{MultiWOZ 2.2 (MW)}
MultiWOZ~\cite{multiwoz} is a widely used DST dataset which follows a standard human-to-human Wizard of Oz approach, spanning several domains. This allows for significantly higher language variety and more complex dialogues, as there are little to no restrictions put on the users when creating data. 
The lack of language restrictions and the \textit{explicit usage of categorical slots} requires inferring values in turns, alongside extractively collecting slot values from utterances. 
An extra challenge is entity bias and misannotations, which have been approached by multiple works~\cite{multiwoz21,multiwoz2.2,multiwoz23,multiwozannotations}.
For training and evaluation, we use the 2.2 variant~\cite{multiwoz2.2}  supported by the original MW authors\footnote{\url{https://github.com/budzianowski/multiwoz}}. MW 2.2 extends the 2.1 version by cleaning some annotations and, not only introducing categorical slot annotations, but also introducing a set of \textit{active user intents} per user turn. We follow the assumption that the \textit{current user intent} is the next to be fulfilled in the active user intent set (i.e. when an intent is removed from the active intent set, the user had been working towards fulfilling it). We use this assumption to retrieve a \textit{single intent} per user utterance.

\vspace{2mm}
\noindent\textbf{Farfetch Simulated Dialogues (Farfetch-Sim).}
This dataset comprises dialogues that simulate a fashion concierge~\cite{ifetch_workshop} that understands customer needs and provides the correct answers. 
These were created in a way that reflects past real user experiences on the Farfetch platform, with a massive number of users.
The simulated dialogues cover the complete customer journey: greeting, product search and exploration, to checkout. Throughout the different conversational journeys, users engage in product-grounded conversations, across different scenarios.
We defined a range of scenarios and flows that reproduce real-world client-assistant interactions and introduce novel fashion-specific sub-dialogues that combine language and product metadata. From a total of 39,956 simulated dialogues, we extract 236,072 annotated utterances (slot-filling and intent) for training, 48,427 for validation and 48,097 testing.

\vspace{2mm}
\noindent\textbf{Farfetch User Dialogues (Farfetch-Costumers).}
This set of real and authentic dialogues was obtained during a user testing session of a  Farfetch's in-house conversational shopping assistant prototype. Users (actual costumers) were sampled based on device (desktop or mobile chat), and clothing gender (men or women), and had no prior experience using a conversational agent for product discovery. 
A total of 85 complete dialogues were annotated with slot-filling and intent detection information, and used for testing. 

\subsection{Training}
Similarly to vanilla BERT-DST, we train the models using randomly sampled batches of size 32. Unless otherwise stated, we used the [BERT base, Uncased] architecture and weights and train for 100 epochs---except for the Farfetch dialogues, which we train for 20 epochs, due to their large amount. We set the learning rate to $2e^{-6}$ and use ADAM optimizer. 

\subsection{Metrics and evaluation methodology}
We evaluate slot-filling using the standard \textbf{joint-goal accuracy} (JG) metric. Joint-goal accuracy is calculated as follows: in dialogue turn $T$, update a set of active slots $S$ (initialized as $\emptyset$ when the dialogue begins) by adding all $(slot\;key,slot\;value)$ pairs present in $T$ so that $S$ contains at most one of each slot keys, replacing ones that were previously present. The joint-goal score for turn $T$ is 1 if $S$ is equal to the ground truth, which is updated in a similar manner. (i.e. \textit{active slots for all current and previous turns have been correctly classified}), and 0 otherwise. The final value is the average of the joint-goal scores of every dialogue turn. The joint-goal score tends to accumulate errors from earlier dialogue turns, unless the system is able to reclassify. We evaluate single-turn dialogues using the slot F1 score, as per JointBERT~\cite{bertjointclassslotfill}.

To evaluate in the M2M dataset, we use the provided BERT-DST \cite{bertdst} evaluation script. In the MultiWOZ dataset, we use the recommended TRADE-DST~\cite{tradedst} pre-processing and evaluation scripts (we refrain from using the special pre-processing considerations for plural nouns). We use different evaluation scripts to ensure that comparisons with other works are adequate. We adapt the TRADE-DST evaluation scripts for the Farfetch dialogues.

\subsection{General Results}
\label{sec:discussion}
In this section we analyze the performance of the proposed approach under different conditions: \textit{no overlap of slots per intent} and \textit{multi-slot per intent}.

\begin{table}[t]
\begin{minipage}{\textwidth}
  \begin{minipage}[b]{0.52\linewidth}
    \centering
    \small
    \begin{tabular}{l|cc|cc}
        \toprule
        & \multicolumn{2}{c|}{\textbf{Sim-M}}   & \multicolumn{2}{c}{\textbf{Sim-R}}\\
        \multicolumn{1}{c|}{\textbf{Model}}     & \textbf{JG}   & \textbf{Int. Acc.} & \textbf{JG} & \textbf{Int. Acc.}   \\   \midrule
        BERT-DST~\cite{bertdst}            &  81.9               & --        & 88.6 & --             \\
        BDST-C                             & 82.6                & --        & 86.1           & --                \\
        BDST-I                             & 83.3       & 100.0     & \textbf{91.3}& 99.9            \\
        \midrule
        TripPy~\cite{tripy} & \textbf{83.5} & -- & 90.0  & -- \\
        \bottomrule
        \end{tabular}
     \centering
     \vspace{3mm}
    \caption{Results on the M2M datasets.
    \label{tab:sim-int-slot-metrics}}
    \end{minipage}
  \hfill
  \begin{minipage}[b]{0.45\linewidth}
    \centering
    \includegraphics[width=\linewidth]{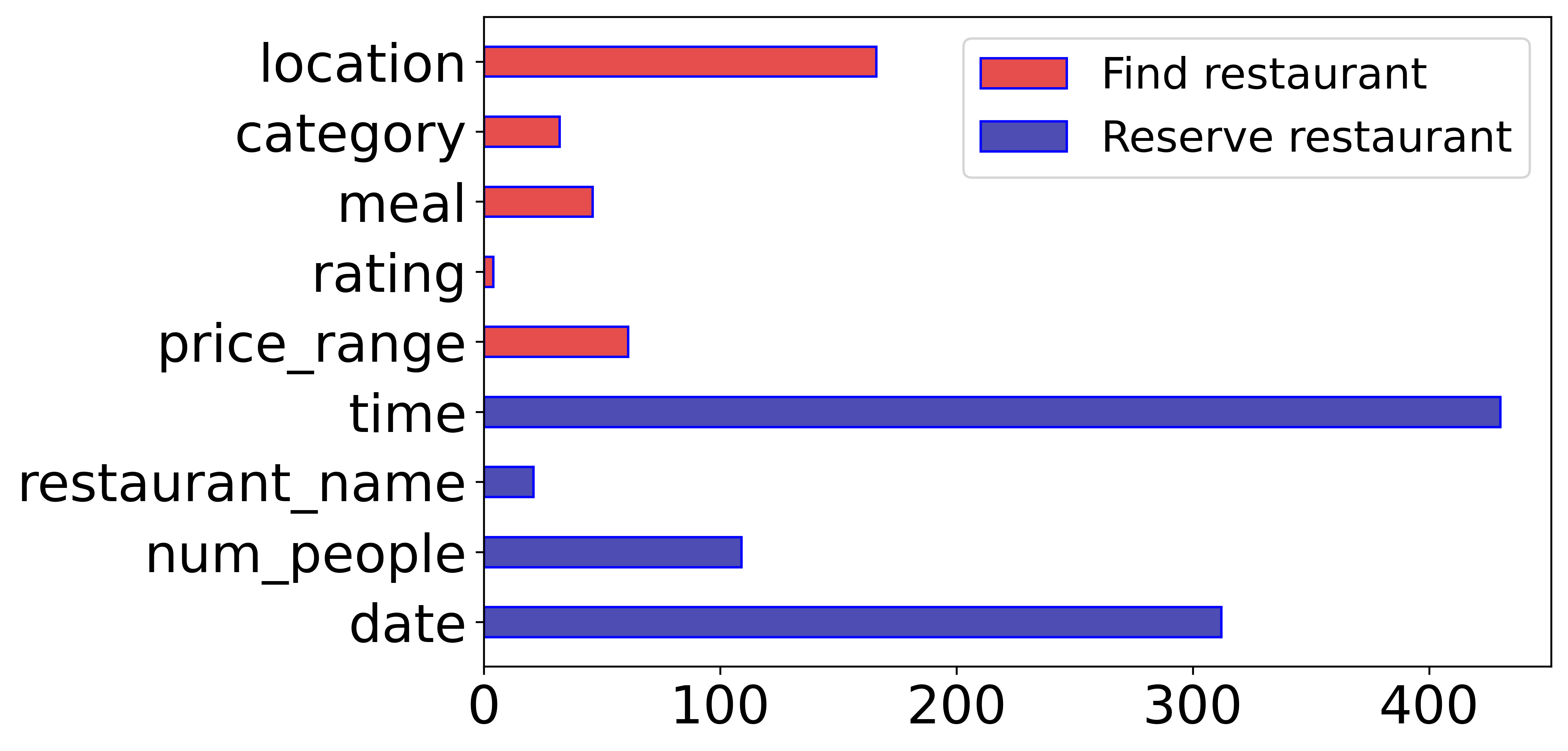}
    \vspace{0.5mm}
    \captionof{figure}{Slot key distribution on the Sim-R train split, by intent.}
    \label{fig:sim-r-slot-intent}
  \end{minipage}
  \vspace{-0.8cm}

\end{minipage}
\end{table}

\subsubsection{No overlap of slots per intent: M2M.}
Table \ref{tab:sim-int-slot-metrics} displays the evaluation metrics on the M2M datasets of our two proposals alongside vanilla BERT-DST performance. To generate an ontology for categorical slots, we use a similar heuristic to the one used for the SGD dataset \cite{schemaguided}: slots which refer to a range of values or a small amount of discrete elements which can easily be listed are categorical, while slots with continuous, uncountable or several values are non-categorical. In Sim-M, we consider the slot \textit{num\_tickets} as categorical---in Sim-R, we consider the \textit{num\_people, price\_range, meal} and \textit{rating} slots. 

The BDST-C performance on Sim-M is quite close to the vanilla model, as expected. This is due to only one slot being considered categorical. It is also important to note that the data was not created with categorical slots in mind---since all slots are explicitly present in dialogue spans, moving away from them may not be ideal for performance; especially relevant in SIM-R. On the other hand, the joint-goal score of BDST-I was higher than anticipated, showing itself to be competitive with the state-of-the-art \cite{tripy}. By analyzing the coarse-grained intent information contained in the data (\textit{none, BUY\_MOVIE\_TICKETS} in Sim-M; \textit{none, FIND\_RESTAURANT, RESERVE\_RESTAURANT} in Sim-R). We find that, in M2M, the user intent directly correlates with the slots that are being mentioned, containing no overlap of mentioned slots, per intent (Figure~\ref{fig:sim-r-slot-intent}). 
The general performance improvement when introducing intent information supports our claim that \textit{jointly training a model on both slot-filling and intent detection tasks can improve performance.}

\setlength\dashlinedash{1pt}
\setlength\dashlinegap{1.5pt}
\setlength\arrayrulewidth{0.5pt}

\begin{table}[t]
  \begin{minipage}{\textwidth}
  \begin{minipage}[b]{0.52\linewidth}
    \centering
    \centering
    \small
    \begin{tabular}{@{}lcc@{}}
        \toprule
        \multirow{2}{*}{\textbf{Model}} & \multicolumn{2}{c}{\textbf{MW 2.2}} \\ 
        & \textbf{JG} & \textbf{Int. Acc.}\\
        \midrule
        BERT-DST~\cite{bertdst}                   & 33.0        & --      \\
        BERT-DST (w/ dialogue history) & 37.6 & --    \\
        \midrule
        BDST-I                    & 40.8        & 88.4         \\
        BDST-C                     & 48.4        & --            \\
        BDST-J                   & 49.0        & 87.9            \\
        \midrule
        BDST-C$^\textit{LARGE}$          & 48.6    & --            \\
        BDST-J$^\textit{LARGE}$         & 49.8          & 87.7 \\
        \midrule
        \multicolumn{3}{c}{\textbf{Systems}}\\
        \midrule
        SGD Baseline~\cite{schemaguided}                                & 42.0*          & -- \\
        TRADE-DST~\cite{tradedst}       & 	45.4*       & -- \\
        DS-DST~\cite{findorclassifydst}                                     & 51.7*      & -- \\
        \bottomrule
    \end{tabular}
    \vspace{3mm}
    \captionof{table}{Joint-goal and intent detection accuracy scores on MultiWOZ 2.2 dataset. Values with * are reported by~\cite{multiwoz2.2}. It should be noted that the DS-DST model uses two BERT models.
    }
    \label{tab:token-additions-metrics-multiwoz}
    \end{minipage}
  \hfill
  \begin{minipage}[b]{0.45\linewidth}
    \centering
    \includegraphics[width=0.9\linewidth]{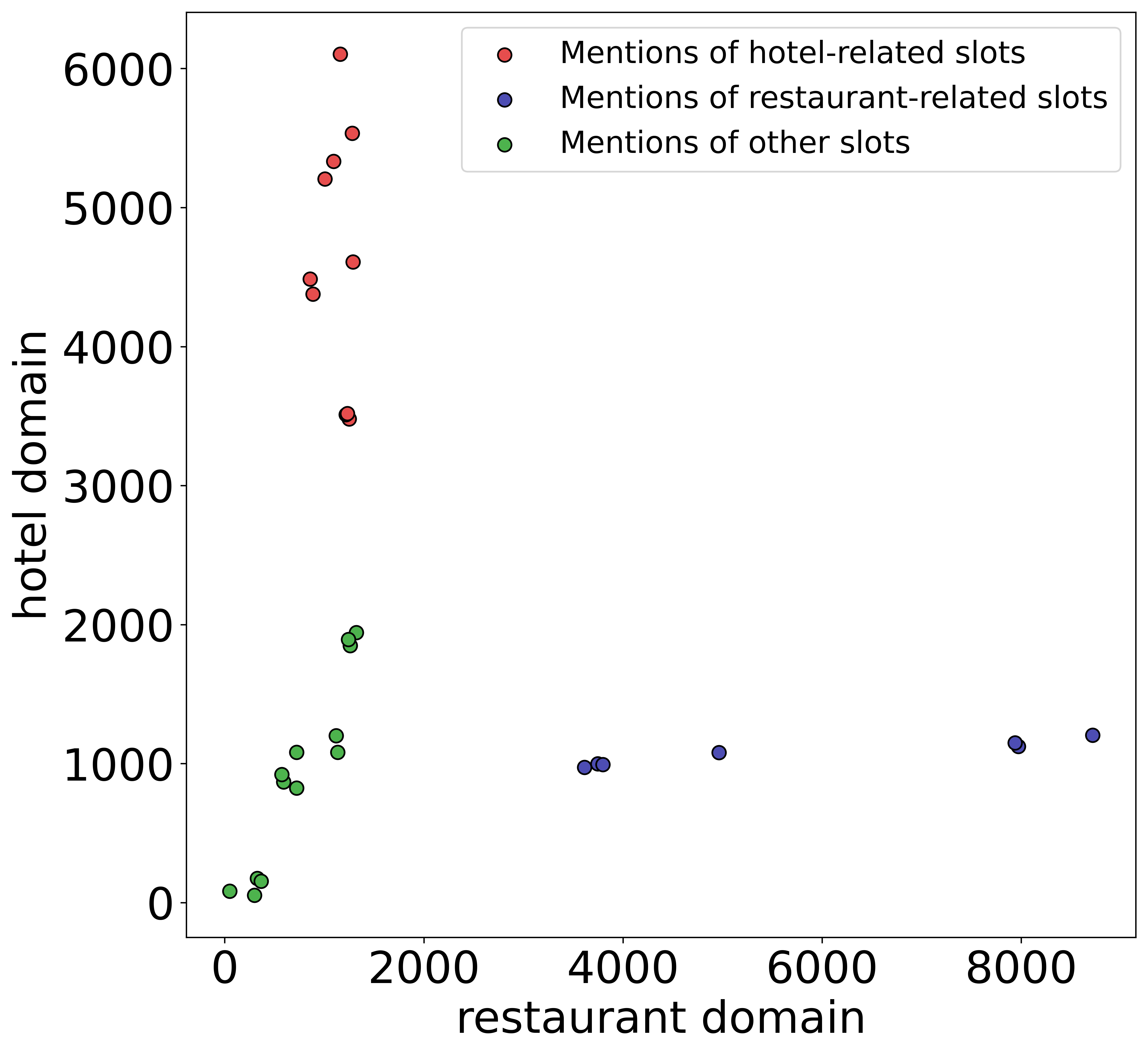}
    \vspace{10mm}
    \captionof{figure}{Cross-domain slot mentions on the MultiWOZ 2.2 in the \textit{hotel} vs. \textit{restaurant} domains.}
    \label{fig:mwoz-slot-intent}
  \end{minipage}
\end{minipage}
\vspace{-8mm}
\end{table}

\subsubsection{Multi-slot per intent: MultiWOZ.}
Leveraged by the insights from the previous experiments and the results on the MultiWOZ dataset (Table \ref{tab:token-additions-metrics-multiwoz}), we reached several conclusions.
First, we observed that training a model for both intent detection and slot-filling improves slot-filling performance. MultiWOZ 2.2, similarly to Sim-R, displays a high correlation between the active intent and the slots that are being mentioned.
Second, the proposed conditioning architecture, i.e. tokens and corresponding heads, enabled our models to approach state-of-the-art performance. 
When compared with TRADE-DST, our model performs significantly better, proving to be a solid alternative for real-world systems where probabilistic outputs are preferred.
Third, introducing more domain information improves overall performance.
The joint-goal score largely increases by simply \textit{introducing categorical slot tokens}. This can be seen when evaluating BERT-DST instances vs. their BDST-C counterparts. A similar result can be seen when introducing intent information---in MultiWOZ, the result of the intent detection task can inform slot-filling modules of the domain relevant to the current utterance. Then,  we show how the domain of the classified user intent is directly related to the frequency of mentioned slots (Figure \ref{fig:mwoz-slot-intent}). When the current domain is restaurant, the \textbf{slot-gate} for hotel related slots is more likely to be correct when outputting \textit{none}, while slot-gates related to restaurant slots are likely to output \textit{span}.
Finally, we also observed that increasing the model size slightly improves performance. In our tests using BERT-large, which contains about 3 times more trainable parameters than BERT-base (345 million vs. 110 million), shows a limited, but consistent, performance gain of less than 1\% in all situations.

\subsubsection{Farfetch Dialogues.} 
Finally, we evaluated the proposed model in an online shopping assistant with both simulated and real costumer dialogues. For this experiment, models are trained solely on simulated dialogues.
Table~\ref{tab:token-additions-metrics-ifetch} reports the obtained results. First, in the simulated dialogues (Farfetch-Sim), we observe that BDST-I can successfully detect both intents and slot-values, with significant improvements in slot F1 and intent accuracy. When we consider dialogues with real costumers, the robustness of BDST-I becomes more evident: the gap in slot-F1, intent accuracy and, more importantly, the joint-goal accuracy between BDST-I and the other two baselines increase considerably. In particular, joint-goal accuracy is 71.0\% and intent accuracy reaches 95.4\%, which confirms that performing both tasks simultaneously and conditionally inferring slot values and intents provides the model with more information to improve its performance.

\begin{table*}[t]
    \centering
    \small
    \begin{tabular}{@{}p{3cm}P{1.5cm}P{1.5cm}P{1.5cm}P{1.5cm}P{1.5cm}@{}}
        \toprule
        \multirow{2}{*}{\textbf{Model}} & \multicolumn{2}{c}{\textbf{Farfetch-Sim}}& \multicolumn{3}{c}{\textbf{Farfetch-Costumers}}\\
        & \textbf{Slot F1} & \textbf{Int. Acc.} & \textbf{Slot F1} & \textbf{Int. Acc.} & \textbf{JG}\\
        \midrule
        JointBERT~\cite{bertjointclassslotfill}           &   93.5  & 96.7 & 83.2 & 93.8 & 54.9\\
        BDST~\cite{bertdst}                &   94.2  & -- & 85.0 & -- & 65.1 \\
        BDST-I               & \textbf{94.6} & \textbf{98.1} & \textbf{87.3} & \textbf{95.4} & \textbf{71.0} \\ 
        \bottomrule
    \end{tabular}
    \vspace{1mm}
    \caption{Joint-goal and intent detection accuracy scores on Farfetch dialogues.}
    \label{tab:token-additions-metrics-ifetch}
    \vspace{-10mm}
\end{table*}

\section{Conclusion}
In the context of this work, we explicitly assumed that there are strong dependencies among language tokens, and that these dependencies become even more salient when the Transformer is conditioned on the dialogue data and on the dialogue state. We proposed an extension to a well-established model, which takes advantage of introducing extra dialogue information and multi-task learning, significantly increasing performance in all cases.
Our contributions are as follows:

\begin{itemize}
    \item \textbf{DST inference task conditioning architecture:} The multi-head architecture and the corresponding tokens elegantly extends the Transformer encoder architecture
    to facilitate joint slot-filling and intent detection. We also observed that training on the different tasks also improved results, thus leveraging the multi-task parameter sharing nature.
    \item \textbf{Multiple slot-filling across domains:} The proposed architecture nicely supports the MultiWOZ 2.2 scenarios where multiple heterogeneous slots co-occur in data, e.g. restaurant span-based slots with hotel categorical slots.
    \item \textbf{State of the art competitive results across heterogeneous domains:} Our models which perform intent detection and slot-filling outperform strong baselines~\cite{tradedst} of equivalent complexity, by learning the intrinsic correlations between the user intent and the slots which are currently being mentioned.
    \item \textbf{Generalization to realistic domain-specific dialogues:} Experiments show that BDST-I effectively generalizes in state-tracking for domain-specific and real scenarios, outperforming the compared approaches.
\end{itemize}
To sum up, we proposed a principled and theoretically well grounded approach to dialogue state tracking that significantly improves performance.
The model is flexible enough be augmented with external heuristics~\cite{tripy}, and generalizes to multiple domains.

\vspace{-1mm}
\section{Acknowledgements}
This work was partially funded by the FCT Scholarship PRT/BD/ 152803 /2021, the NOVA LINCS project (UIDP/04516/2020), and the CMU Portugal project iFetch (LISBOA-01-0247-FEDER-045920). 

\bibliographystyle{splncs04}
\bibliography{main}

\begin{thebibliography}{10}
\providecommand{\url}[1]{\texttt{#1}}
\providecommand{\urlprefix}{URL }
\providecommand{\doi}[1]{https://doi.org/#1}

\bibitem{coefficients}
Akoğlu, H.: User's guide to correlation coefficients. Turkish Journal of
  Emergency Medicine  \textbf{18},  91 -- 93 (2018)

\bibitem{multiwoz}
Budzianowski, P., Wen, T.H., Tseng, B.H., Casanueva, I., Stefan, U., Osman, R.,
  Ga{\v{s}}i\'c, M.: Multiwoz - a large-scale multi-domain wizard-of-oz dataset
  for task-oriented dialogue modelling. In: EMNLP (2018)

\bibitem{bertdst}
Chao, G.L., Lane, I.: {BERT-DST}: Scalable end-to-end dialogue state tracking
  with bidirectional encoder representations from transformer. In: INTERSPEECH
  (2019)

\bibitem{bertjointclassslotfill}
Chen, Q., Zhuo, Z., Wang, W.: Bert for joint intent classification and slot
  filling. ArXiv  \textbf{abs/1902.10909} (2019)

\bibitem{bertlookat}
Clark, K., Khandelwal, U., Levy, O., Manning, C.D.: What does {BERT} look at?
  an analysis of {BERT}{'}s attention. In: Proceedings of the 2019 ACL Workshop
  BlackboxNLP: Analyzing and Interpreting Neural Networks for NLP. pp. 276--286
  (Aug 2019)

\bibitem{snips}
Coucke, A., Saade, A., Ball, A., Bluche, T., Caulier, A., Leroy, D., Doumouro,
  C., Gisselbrecht, T., Caltagirone, F., Lavril, T., et~al.: Snips voice
  platform: an embedded spoken language understanding system for
  private-by-design voice interfaces. arXiv preprint arXiv:1805.10190 pp.
  12--16 (2018)

\bibitem{devlin-etal-bert}
Devlin, J., Chang, M.W., Lee, K., Toutanova, K.: {BERT}: Pre-training of deep
  bidirectional transformers for language understanding. In: Proceedings of the
  2019 Conference of the North {A}merican Chapter of the Association for
  Computational Linguistics: Human Language Technologies, Volume 1 (Long and
  Short Papers). pp. 4171--4186 (Jun 2019)

\bibitem{multiwoz21}
Eric, M., Goel, R., Paul, S., Sethi, A., Agarwal, S., Gao, S., Kumar, A.,
  Goyal, A.K., Ku, P., Hakkani{-}T{\"{u}}r, D.: Multiwoz 2.1: {A} consolidated
  multi-domain dialogue dataset with state corrections and state tracking
  baselines. In: Calzolari, N., B{\'{e}}chet, F., Blache, P., Choukri, K.,
  Cieri, C., Declerck, T., Goggi, S., Isahara, H., Maegaard, B., Mariani, J.,
  Mazo, H., Moreno, A., Odijk, J., Piperidis, S. (eds.) Proceedings of The 12th
  Language Resources and Evaluation Conference, {LREC} 2020, Marseille, France,
  May 11-16, 2020. pp. 422--428. European Language Resources Association
  (2020), \url{https://aclanthology.org/2020.lrec-1.53/}

\bibitem{atis}
Hakkani-Tur, D., Tur, G., Celikyilmaz, A., Chen, Y.N., Gao, J., Deng, L., Wang,
  Y.Y.: Multi-domain joint semantic frame parsing using bi-directional
  rnn-lstm. In: Proceedings of Interspeech (2016)

\bibitem{multiwoz23}
Han, T., Liu, X., Takanobu, R., Lian, Y., Huang, C., Wan, D., Peng, W., Huang,
  M.: Multiwoz 2.3: A multi-domain task-oriented dialogue dataset enhanced with
  annotation corrections and co-reference annotation. In: Proceedings of the
  10th CCF International Conference on Natural Language Processing and Chinese
  Computing. pp. 206--218. CCF (2021)

\bibitem{tripy}
Heck, M., van Niekerk, C., Lubis, N., Geishauser, C., Lin, H., Moresi, M.,
  Gasic, M.: Trippy: {A} triple copy strategy for value independent neural
  dialog state tracking. In: Pietquin, O., Muresan, S., Chen, V., Kennington,
  C., Vandyke, D., Dethlefs, N., Inoue, K., Ekstedt, E., Ultes, S. (eds.)
  Proceedings of the 21th Annual Meeting of the Special Interest Group on
  Discourse and Dialogue, SIGdial 2020, 1st virtual meeting, July 1-3, 2020.
  pp. 35--44. Association for Computational Linguistics (2020),
  \url{https://aclanthology.org/2020.sigdial-1.4/}

\bibitem{simpletod}
Hosseini-Asl, E., McCann, B., Wu, C.S., Yavuz, S., Socher, R.: A simple
  language model for task-oriented dialogue. NeurIPS  \textbf{2020-December} (5
  2020)

\bibitem{shoptalk}
Manku, G., Lee{-}Thorp, J., Kanagal, B., Ainslie, J., Feng, J., Pearson, Z.,
  Anjorin, E., Gandhe, S., Eckstein, I., Rosswog, J., Sanghai, S., Pohl, M.,
  Adams, L., Sivakumar, D.: Shoptalk: {A} system for conversational faceted
  search. CoRR  \textbf{abs/2109.00702} (2021),
  \url{https://arxiv.org/abs/2109.00702}

\bibitem{improvingslotfilling}
Pouran Ben~Veyseh, A., Dernoncourt, F., Nguyen, T.H.: Improving slot filling by
  utilizing contextual information. In: Proceedings of the 2nd Workshop on
  Natural Language Processing for Conversational AI. pp. 90--95 (Jul 2020)

\bibitem{multiwozannotations}
Qian, K., Beirami, A., Lin, Z., De, A., Geramifard, A., Yu, Z., Sankar, C.:
  Annotation inconsistency and entity bias in {M}ulti{WOZ}. In: Proceedings of
  the 22nd Annual Meeting of the Special Interest Group on Discourse and
  Dialogue. pp. 326--337 (Jul 2021)

\bibitem{stackpropagation}
Qin, L., Che, W., Li, Y., Wen, H., Liu, T.: A stack-propagation framework with
  token-level intent detection for spoken language understanding. In:
  EMNLP-IJCNLP. pp. 2078--2087 (Nov 2019)

\bibitem{schemaguided}
Rastogi, A., Zang, X., Sunkara, S., Gupta, R., Khaitan, P.: Towards scalable
  multi-domain conversational agents: The schema-guided dialogue dataset. In:
  AAAI (2020)

\bibitem{SIM}
Shah, P., Hakkani-T{\"u}r, D., Liu, B., T{\"u}r, G.: Bootstrapping a neural
  conversational agent with dialogue self-play, crowdsourcing and on-line
  reinforcement learning. In: NAACL. pp. 41--51 (Jun 2018)

\bibitem{ifetch_workshop}
Sousa, R.G., Ferreira, P.M., Costa, P.M., Azevedo, P., Costeira, J.P.,
  Santiago, C., Magalhaes, J., Semedo, D., Ferreira, R., Rudnicky, A.I.,
  Hauptmann, A.G.: Ifetch: Multimodal conversational agents for the online
  fashion marketplace. In: Proceedings of the 2nd ACM Multimedia Workshop on
  Multimodal Conversational AI. p. 25–26. MuCAI'21, Association for Computing
  Machinery, New York, NY, USA (2021). \doi{10.1145/3475959.3485395},
  \url{https://doi.org/10.1145/3475959.3485395}

\bibitem{todbert}
Wu, C.S., Hoi, S.C., Socher, R., Xiong, C.: {TOD}-{BERT}: Pre-trained natural
  language understanding for task-oriented dialogue. In: EMNLP. pp. 917--929
  (Nov 2020)

\bibitem{tradedst}
Wu, C.S., Madotto, A., Hosseini-Asl, E., Xiong, C., Socher, R., Fung, P.:
  Transferable multi-domain state generator for task-oriented dialogue systems.
  In: Proceedings of the 57th Annual Meeting of the Association for
  Computational Linguistics. pp. 808--819 (Jul 2019)

\bibitem{wu-etal-2020-slotrefine}
Wu, D., Ding, L., Lu, F., Xie, J.: {S}lot{R}efine: A fast non-autoregressive
  model for joint intent detection and slot filling. In: EMNLP. pp. 1932--1937
  (Nov 2020)

\bibitem{unknownslotvaluesdst}
Xu, P., Hu, Q.: An end-to-end approach for handling unknown slot values in
  dialogue state tracking. In: Proceedings of the 56th Annual Meeting of the
  Association for Computational Linguistics (Volume 1: Long Papers). pp.
  1448--1457 (Jul 2018)

\bibitem{multiwoz2.2}
Zang, X., Rastogi, A., Sunkara, S., Gupta, R., Zhang, J., Chen, J.: Multiwoz
  2.2: A dialogue dataset with additional annotation corrections and state
  tracking baselines. In: Proceedings of the 2nd Workshop on Natural Language
  Processing for Conversational AI, ACL 2020. pp. 109--117 (2020)

\bibitem{purelytransformerdst}
Zeng, Y., Nie, J.Y.: Multi-domain dialogue state tracking -- a purely
  transformer-based generative approach (2020)

\bibitem{findorclassifydst}
Zhang, J., Hashimoto, K., Wu, C., Wan, Y., Yu, P.S., Socher, R., Xiong, C.:
  Find or classify? dual strategy for slot-value predictions on multi-domain
  dialog state tracking. CoRR  \textbf{abs/1910.03544} (2019),
  \url{http://arxiv.org/abs/1910.03544}

\end{thebibliography}

\end{document}